\documentclass[conference]{IEEEtran}
\IEEEoverridecommandlockouts

\usepackage[bookmarks=false]{hyperref}
\usepackage{booktabs}
\usepackage{float}
\usepackage{graphicx}
\usepackage{caption}
\usepackage{subcaption}
\usepackage{cite}
\usepackage{amsmath,amssymb,amsfonts}
\usepackage{algorithmic}
\usepackage{graphicx}
\usepackage{textcomp}
\usepackage{xcolor}
\usepackage[colorinlistoftodos]{todonotes}

\floatstyle{plaintop}
\restylefloat{table}

\def\BibTeX{{\rm B\kern-.05em{\sc i\kern-.025em b}\kern-.08em
    T\kern-.1667em\lower.7ex\hbox{E}\kern-.125emX}}
\begin{document}

\title{Time to Die: Death Prediction in Dota 2 using Deep Learning
\thanks{This work was supported by the EPSRC Centre for Doctoral Training in Intelligent Games \& Games Intelligence (IGGI) [EP/L015846/1] and the Digital Creativity Labs (digitalcreativity.ac.uk), jointly funded by EPSRC/AHRC/Innovate UK under grant no. EP/M023265/1.}}

\author{
\IEEEauthorblockN{Adam Katona\IEEEauthorrefmark{1}, Ryan Spick\IEEEauthorrefmark{1}, Victoria J. Hodge\IEEEauthorrefmark{1}, Simon Demediuk\IEEEauthorrefmark{1},\\ Florian Block\IEEEauthorrefmark{2}, Anders Drachen\IEEEauthorrefmark{1} and James Alfred Walker\IEEEauthorrefmark{1},~\IEEEmembership{Senior Member, IEEE}}\\

\IEEEauthorblockA{\IEEEauthorrefmark{1}Department of Computer Science, University of York, Heslington, York, YO10 5GE, UK}

\IEEEauthorblockA{\IEEEauthorrefmark{2}Department of Theatre, Film and Television, University of York, Heslington, York, YO10 5GE, UK\\
Email: \{ak1774, rjs623, victoria.hodge, simon.demediuk, florian.block, anders.drachen, james.walker\}@york.ac.uk}
}

%

\IEEEpubid{\begin{minipage}{\textwidth}\ \\[12pt]
978-1-7281-1884-0/19/\$31.00 \copyright 2019 IEEE
\end{minipage}}

\maketitle

\begin{abstract}

Esports have become major international sports with hundreds of millions of spectators. Esports games generate massive amounts of telemetry data. Using these to predict the outcome of esports matches has received considerable attention, but micro-predictions, which seek to predict events inside a match, is as yet unknown territory. Micro-predictions are however of perennial interest across esports commentators and audience, because they provide the ability to observe events that might otherwise be missed: esports games are highly complex with fast-moving action where the balance of a game can change in the span of seconds, and where events can happen in multiple areas of the playing field at the same time. Such events can happen rapidly, and it is easy for commentators and viewers alike to miss an event and only observe the following impact of events. In Dota 2, a player hero being killed by the opposing team is a key event of interest to commentators and audience. We present a deep learning network with shared weights which provides accurate death predictions within a five-second window. The network is trained on a vast selection of Dota 2 gameplay features and professional/semi-professional level match dataset. Even though death events are rare within a game (1\% of the data), the model achieves 0.377 precision with 0.725 recall on test data when prompted to predict which of \textit{any} of the 10 players of either team will die within 5 seconds. An example of the system applied to a Dota 2 match is presented. This model enables real-time micro-predictions of kills in Dota 2, one of the most played esports titles in the world, giving commentators and viewers time to move their attention to these key events.





\end{abstract}


\section{Introduction}


The term esports describes video games that are played competitively and usually watched by large audiences~\cite{hamari2017esports}. Superdata~\cite{superdata:17} predict 330 million esports spectators by 2019.  Esports have become an important research field across academia and industry~\cite{schubert2016esports} due to the availability of high-dimensional, high-volume data from virtually every match. This has introduced the field of esports analytics: ``\textit{the process of using esports related data, [...], to find meaningful patterns and trends in said data, and the communication of these patterns using visualization techniques to assist with decision-making processes}''~\cite{schubert2016esports}. This definition highlights a fundamental challenge in esports: making the matches comprehensible to the audience. Many esports are complex and fast-paced, making it hard to fully unpack the live action with the naked eye.


Esports analytics has to a significant extent focused on the Multi-player Online Battle Arena (MOBA) genre~\cite{schubert2016esports}. As a typical MOBA, \textit{Dota 2} by Valve Corp. is a 10 player game with two teams of five players (taking typically between 30-40 minutes~\cite{florianspaper}). Dota 2 is a very complex game where each player selects one of many unique 'heroes' to play. Each player uses their hero's spells and abilities coupled with items they purchase in-game to gather resources across the 3-lane map. The heroes play different roles where they aim to generate resources via fights against the rival team to progress through hero levels and become more powerful. Each team's ultimate aim is to destroy the opposing team's base. There are multiple strategies and winning a game requires coordination within the team and the ability to react to the opposition's tactics and behavior. The game is real-time with hidden information and has deep strategic gameplay. 


During a Dota 2 game, the players must closely monitor their hero's status, in particular, the likelihood of dying. Thus, a method of predicting deaths within a game of Dota 2 can be beneficial on multiple levels of both professional and amateur game analysis. In current esports analytics approaches, many of the performance metrics (data variables) used are correlated directly to team success rather than necessarily to a player's likelihood to die.  In particular, in Section~\ref{sec:Results}, we show that the hero health variable which may seem an obvious indicator of likelihood to die is not correlated to hero deaths. Some heroes have abilities which allow them to heal themselves or their team-mates while heroes can purchase items in-game that allow them to heal or teleport away from danger. Hence, we cannot simply use player health for predicting deaths as the prediction is complex and requires careful investigation. 
For the performance of players and heroes to be accurately analysed, the historical data needs to be carefully considered and tailored to the specific task. Applying performance metrics without careful selection introduces noise and leads to biased or skewed analyses.




This paper introduces the first step towards the larger challenge of \emph{micro-predictions} in esports. Micro-predictions are granular predictions about what might happen in the near future in the game, as opposed to predicting the outcome of the match itself. The concept is known from other sports, e.g. predicting which football player that will score the next goal. Understanding the likelihood of dying in the near future can be vital in determining what tactics a player should use within post-game analysis. The use of a prediction tool within the professional game broadcast scene would provide far more readability for the audience on why and how players are making mistakes and what tactics they should use individually and as a team, allowing commentators to simplify complex build ups to when a player dies. Live broadcasts would benefit from feeding this information back to the audience, alongside automated camera movement to direct the focus when these predictions occur to a high enough degree of accuracy, making both the commentator's job and the users' viewing experience a far smoother process. Furthermore, post-game data analysis forms a large part of professional esport players' training schedule~\cite{block2017develop}. Coaches currently have to select (by hand) points of the game that led to deaths or where a team's advantage snowballed. With access to our tool, automated points could be created where the model would predict a player's death, whether they did die or not allowing a further investigation as to why the model assumed they were in danger. We introduce a novel prediction framework using a shared memory neural network (described in Section~\ref{sec:Model}) and analyse a much larger set of in-game features compared to previous esports work.


The experimental design behind this paper was to train a neural network model to predict whether a hero will die in the next 5 seconds. A 5-second window provides a large enough period before a kill occurs, such that commentators would be able to make use of any prediction information the network provides. We ran preliminary experiments exploring larger window sizes from 5 seconds up to 20 seconds, with the precision of the model drastically dropping as the window size increased. 

There are certain immediate limitations, being that every data point is required, even that of unseen hero data. With this in mind, the tool would be limited to post-game analyses, or in a developer/broadcaster environment where the required variables can be obtained (an extension of a professional game broadcast for example).
Another limitation is that the game is constantly changing, meaning our model might not generalize well to newer game versions.

The structure of the paper is as follows; Related work will be discussed in Section~\ref{sec:relatedwork}. Followed by an in-depth analysis of the data set, including the cleaning and feature selection based on domain knowledge in Section~\ref{sec:dataset}. In Section~\ref{sec:Model} we then explore the neural network model used, from dealing with the imbalance data, to in-depth hyperparameter exploration.
We then summarise results of the fully trained model and how it applies to the test data and a selected Dota 2 match in real time in Section~\ref{sec:Results}. Section~\ref{sec:discussion} discusses the difficulties of performance elevation. We provide concluding remarks in Section~\ref{sec:conclusion}. The source code used for creating the dataset and training the model is available at \url{https://github.com/adam-katona/dota2\_death\_prediction}.


\section{Related Work}\label{sec:relatedwork}


Collectively, the esports industry is generating knowledge at a rapidly increasing pace. However, much of this knowledge is not publicly available due to commercial confidentiality, so establishing the state-of-the-art in esports is challenging. Research into esports has previously been published across a number of different disciplines, these include; AI, analytics, psychology, education, visualization, ethnography, marketing, management and business (e.g. \cite{Yannkakis2012-GAI,yang2014identifying,schubert2016esports,seo2013,hamari2017esports,xue2016}).

Within esports, ``esports analytics'' is generally used to denote business intelligence work centred on esports, i.e., the analysis of data such as behavioural telemetry, sales data, etc.. These analyses occur across the range of disciplines and describe a broad area of work. Our work focuses on the analysis of behavioural telemetry data from esports games.


The majority of academic work in this specific area has  focused on building machine learning models to predict the outcome of esports matches or other prediction tasks~\cite{LoLsurvivalanalysis, wang2016predicting, yang2014identifying, hodge2018:arXiv, xue2016,2017arXiv170103162Y, synnaeve2011bayesian, bursztein2016legend}. Other work includes Schubert et al.~\cite{schubert2016esports}, who described a method for detecting spatio-temporally bounded team encounters in Dota 2. Summerville et al.~\cite{summerville2016draft} used machine learning to predict draft picks in Dota 2.  Rioult~\cite{rioult2014mining} discussed some of the general applications of mining player tracks from esports games. Drachen et al.~\cite{drachen2014skillbased} used classification for studying the movement patterns of Dota 2 teams across skill levels. Gao et al.~\cite{Gao2013} classified Dota 2 heroes based on performance metrics. Eggert et al.~\cite{eggert2015classification} used logistic regression to classify players into pre-determined roles using performance metrics. 

The most closely related work is Cleghern et al.~\cite{Cleghern2017:FDG} who predicted hero health in Dota 2 using a combination of techniques: an auto-regressive moving average model to predict small changes in health and a statistical estimation model (see~\cite{Cleghern2017:FDG}) which predicts large changes in health and works in conjunction with logistic and linear regression that predict the sign and magnitude of that change respectively. 
The authors note a peak accuracy of 77.2\% for the ARMA model to predict small changes in health (less than 100 points) five seconds into the future. The statistical estimation model predicts large health changes 10 seconds into the future. However, the model has poor performance in terms of predicting the points of major health changes, although when it does find them, the accuracy is around 80\% for the direction and magnitude of the health change. The models of Cleghern et al.~\cite{Cleghern2017:FDG} only use health data, unlike the present work which uses a large number of features. Furthermore, their dataset comprises just 542 matches, and the level of the matches is not described in the paper, so it is unknown whether the results would work for professional matches and/or amateur levels. 


In summary, while much previous work focuses on macro-predictions (predicting the winning team), there has been a small amount of work on micro-level events like encounters and hero health changes~\cite{Cleghern2017:FDG,drachen2014skillbased,eggert2015classification,rioult2014mining,schubert2016esports} but only one previous work has focused on in-game forecasting with mixed success. None of the existing work has focused on professional/semi-professional levels or predicted hero deaths. Death events are an important part of the game narrative and equally important for broadcasters' commentaries, spectators' understanding of the game state and players' tactics and analyses. From the state-of-the-art, we can further conclude that esports analytics is complex, requires careful consideration of the data and models used and, frequently, no one technique excels.


\section{Dataset}\label{sec:dataset}


Access to large amounts of data is required in order to successfully train a classification model to predict kills. To generate this we use Dota 2 replay files; these are binary files produced by Valve Corp. for every Dota 2 match played.

They contain all low-level game events from the match. They are a comprehensive representation of the matches and are used by Dota 2 engines to recreate entire matches for re-watching and analysis, so are an ideal source of data to train our models. We explain how we obtain these data and describe the format in Section~\ref{sec:DataCollection} and \ref{sec:DataProcessing}, respectively.

\begin{table*}[ht]
\centering
\small
\caption{Details of different categories of features.}
\label{table:featureset}
\begin{tabular}{@{}lllll@{}}
\toprule
Hero statistics & Hero state & Positions / Proximities & Ability & Items list \\ \midrule
\scriptsize
\begin{tabular}[1]{@{}l@{}}FirstBloodClaimed\\ TeamFightParticipation\\ Level\\ Kills\\ Deaths\\ Assists\\ ObserverWardsPlaced\\ SentryWardsPlaced\\ CreepsStacked\\ CampsStacked\\ RunePickups\\ TowerKills\\ RoshanKills\\ TotalEarnedGold\\ LastHitCount\\ TotalEarnedXP\\ Stuns\\ \\ \\ \\ \\\end{tabular} & \scriptsize
\begin{tabular}[1]{@{}l@{}}Agility\\ AgilityTotal\\ Intellect\\ IntellectTotal\\ Strength\\ StrengthTotal\\ MagicalResistanceValue\\ PhysicalArmorValue\\ Mana\\ MaxMana\\ TauntCooldown\\ BKBChargesUsed\\ AbilityPoints\\ PrimaryAttribute\\ MoveSpeed\\ Health\\ MaxHealth\\ DamageMax\\ DamageMin\\ lifeState\\ TaggedAsVisibleByTeam\end{tabular} & 
\scriptsize
\begin{tabular}[1]{@{}l@{}}Hero position x\\ Hero position x change\\ Hero position y\\ Hero position y change\\ Ally proximity 1-4\\ Ally proximity 1-4 change\\ Enemy proximity 1-5\\ Enemy proximity 1-5 change\\ Closest ally tower proximity\\ Closest ally tower proximity change\\ Closest enemy tower proximity\\ Closest enemy tower proximity change\\ \\ \\ \\ \\ \\ \\ \\ \\ \\  \end{tabular} &
\scriptsize
\begin{tabular}[1]{@{}l@{}}Level\\ CastRange\\ ManaCost\\ Cooldown\\ Activated\\ ToggleState\\ \\ \\ \\ \\ \\ \\ \\ \\ \\ \\ \\ \\ \\ \\ \\ \end{tabular} & 
\scriptsize
\begin{tabular}[1]{@{}l@{}}Blink dagger\\ Black king bar\\ Magic wand\\ Quelling blade\\ Power treads\\ Hand of midas\\ Hurricane pike\\ Force staff\\ Abyssal blade\\ Mask of madness\\ Nullifier\\ Travel boots\\ Dagon 5\\ Lotus orb\\ Tpscroll\\ Smoke of deceit\\ Clarity\\ \\ \\ \\ \\ \end{tabular} \\ \bottomrule
\end{tabular}
\end{table*}

\subsection{Data collection}\label{sec:DataCollection}


To allow us to access the replay files, we need the URL of each file. OpenDota~\cite{steam} provides an API for accessing Dota 2 replay URLs that then allows us to download matches from Valve's servers. We collected the last 5000 Professional (major tournaments) and 5000 Semi-Professional (minor tournaments and leagues) games that were played prior to the $5^{th}$ of December 2018. Accounting for errors in downloaded game data, this resulted in a total collection of 9883 matches. To extract the data from the binary files, we used the open source parser Clarity~\cite{clarity}. Clarity provides functions to traverse the game object hierarchy and to access the attributes of game objects. 

To turn the replay into a time-series, we recorded values of a set of attributes (see next section for details) with a given sampling period. Neighboring data points are highly correlated, thus lowering the sampling period results in more data with less diversity. To get most out of our replay files while keeping the size of the dataset manageable, we chose the sampling period to be 4 game ticks, which correspond to 0.133 seconds in game time. To calculate the exact time of death we used the full resolution data however, which resulted in more accurate classification labels. There were some missing key feature values within a number of replay files, for example, the attributes about hero statistics. After discarding these files 7311 replay files remained. To the best of our knowledge, there was no bias introduced by discarding these files.

\subsection{Data processing}\label{sec:DataProcessing}
The data processing stage included cleaning, creating additional features, calculating labels and normalization. 

In order to keep the time series data consistent in the game, we removed all ticks that occurred during a pause in the game. In a Dota2 game, a pause can be initiated by any player, this occurs when technical faults happen within the game, giving referees a chance to fix any issues. Since predicting death during a paused game is trivial, any pause data points were removed from the dataset.



Due to the nature of the predictions we are trying to achieve, i.e. death prediction, we needed to build a feature set that would capture the relevant data. We began by looking at the two most important factors that result in a player's death, relative strength of the heroes and their current location on the map at a given time. Building from an initially small number of features to represent this we ended with a final feature set of 287 features per hero. Some of these features were the values of game object attributes like hero health. Other features were derived values like proximity to other players. Below an overview is given for each category of features, and why they were included. Further details are contained in Table~\ref{table:featureset}. Due to the number of features and the page limit, a full description of each feature is not given, but is available on request by contacting the first author.
 
\begin{itemize}
\item Time (1 feature): The game time might be a useful feature for two reasons. One is that there are events which happen on given times (creature spawns), the other is that as time progresses, the players generally get more powerful, which results in the players pursuing different objectives. This gameplay change is often referred to as transitioning from early-game to mid-game and then late-game.
\item Current state of the hero (21 features): Includes attributes like health, strength, agility, etc. Full list in Table \ref{table:featureset}.
\item Statistics about the hero (17 features): Describes what the hero did in the past. For example the number of kills, deaths, last hits, experience points etc. Full list in Table \ref{table:featureset}.
\item Activatable items (34 features): Players can buy items which can be activated to provide a powerful effect. These items can have a drastic effect on the outcome of an encounter, for example, they can make the player teleport away from danger. Each item has a binary feature value representing whether the player has the item or not, and a cooldown feature, which represents the time until it can be activated again. To keep the feature space low, the 17 most powerful items were selected based on expert opinion. A full list is given in Table \ref{table:featureset}
\item Hero abilities (48 features): Include the attributes of hero abilities like cooldown, level, mana cost, etc. (full list in Table \ref{table:featureset}). For heroes with fewer than eight abilities, values are zero padded.
\item Hero ID (130 features): The identity of the hero is represented as a one hot encoded vector. This feature provides useful information since different heroes have different strengths and weaknesses. However, since the number of heroes is high, this represents a significant portion of the feature space.
\item Hero position, position change (4 features). The position is represented as x and y coordinates. 
\item Hero proximity, proximity change (18 features): Represent the distance between the current player to every other player, and the rate of change of the distances.
\item Distance to closest alive enemy and ally tower, and distance change (4 features): Towers are destroyable structures which can deal a large amount of damage.
\item Visibility history (10 features): Binary features representing whether the player was visible for the enemy team in the past 10 seconds with 1-second resolution. 
\end{itemize}

The game is fully observable, the replay file contains the whole game state.  At first glance adding features depending on the past might seem unnecessary. However an important part of the system are the players, and the behavior of the players does depend on the past. For example, if an enemy just went out of sight, the player still knows that the enemy is in the area. On the other hand, if the enemy disappeared minutes ago, that enemy can be anywhere from the player's perspective. This was our motivation for adding the visibility history features.

Even though a lot of information is present in these features, there is still important information missing which a human observer pays attention to. For example, there is no information about projectiles or non-player creatures.

The classification label is a binary number representing whether or not the player will die within 5 seconds. 

Normalizing the features was done across players. For each feature, we calculated the maximum and minimum values that any player had in any match, and used these values to scale the features to the range between zero and one.

\section{The Model}\label{sec:Model} 

\begin{figure*}[t]
\centering
  \includegraphics[width=.80\linewidth]{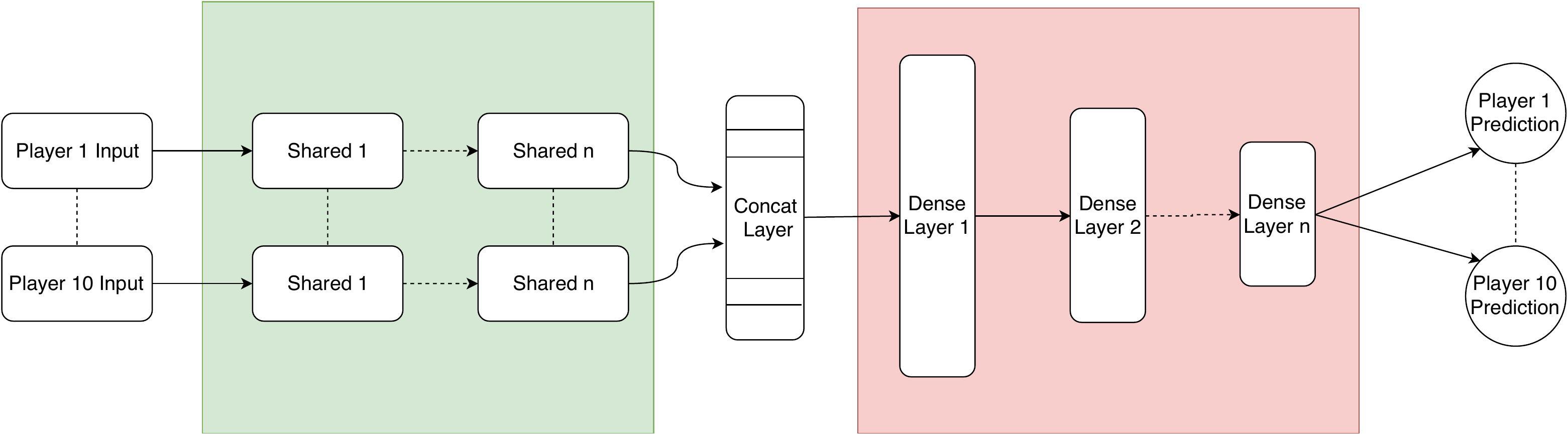}
\caption{Overview of the network architecture. The features from each hero flow through the shared layers, which calculate the same representation for each hero. These representations are concatenated to one big vector, which is passed to a standard fully connected network. The network has 10 outputs, each corresponding to the probability of each hero dying within 5 seconds. }
\label{fig:network}
\end{figure*}

We used a deep, feedforward neural network with weight sharing as our model~\cite{goodfellow2016deep,lecun1995convolutional}. The input of the neural network consists of the features (discussed in the \ref{sec:DataProcessing}) for each hero. In the largest feature set, a hero has 287 features and there are 10 heroes so the whole network has 2870 inputs. The network has 10 independent continuous outputs representing the probability of each player dying in the next 5 seconds. In this section, we first discuss weight sharing and our network architecture. Then we present our methods for data balancing, the training procedure and the hyperparameter exploration.

\subsection{Weight sharing}


Weight sharing can drastically reduce the number of weights of a network without sacrificing much of the representational power~\cite{lecun1995convolutional}. This can be achieved by using the inherent symmetries of the domain. In our case, the symmetry comes from all heroes having exactly the same kind of features, and the hero slot order being irrelevant to the gameplay. 
Weight sharing has been applied to various domains; from image~\cite{Lecun98gradient-basedlearning} and audio~\cite{lee2009unsupervised} classification to natural language processing~\cite{huang2013cross}. One of the most successful and best-known example of weight sharing is the convolutional neural network~\cite{Lecun98gradient-basedlearning}. In the case of image recognition, the goal is to learn positional invariant representations, since we want to recognize objects which can be anywhere in the image. In the prediction problem discussed within this paper, we want to learn hero slot invariant representations.

\subsection{Network architecture}

Fig.~\ref{fig:network} shows a brief abstract overview of the network structure, though the number of layers and depth of the network were not shown. Instead, these values can be seen within Table~\ref{table:hyperparameters}, which also shows the hyperparameters of the network. 
The main takeaway from the network Fig.~\ref{fig:network}, visualises how the inputs entered into the shared network structure.

Here all the weights are shared, essentially creating a sub-network that learns a new, denser representation for the hero features. The concatenation layer appends the outputs from each shared layer creating a tensor of inputs to the last part of the network structure, which is a fully-connected network.

Rectified linear unit (RELU)~\cite{goodfellow2016deep} was used for the output of every hidden layer, while we used the Sigmoid function~\cite{goodfellow2016deep} for the final activation of the network.


\subsection{Dealing with imbalanced data}

The original data-set in its continuous form is extremely in-balanced. Since most of the time during a match the players are not about to die there are a significant amount more points for training on ''no death" labels, with only around 1 \% of the classification data being positive ''death" labels. To balance the training data, we utilised under sampling~\cite{chawla2004special} on the negative labels.  This process was split into two steps;





After parsing the replays files into their readable continuous format, we randomly discarded around 50\% of the negative labels from the data set (where no player died for at least 5 seconds). This saved us a considerable amount of storage space, and made loading the dataset far faster.



The second step of balancing the data took place when we assembled our minibatch for gradient descent \cite{goodfellow2016deep}. Since when someone dies, others do not necessarily die around the same time it is not possible to get batches where the labels are balanced for each player. Instead, we balanced the batch for only one player. At each iteration, a player was selected randomly and the dataset was sampled in a way that the number of negative and positive labels were equal for this player. Then we only backpropagated the error for the selected player.

\subsection{Feature sets}

Because of the complexity of the game, we were not sure if the network can make use of all the feature categories. To gain insight we created three separate feature sets, and run three training procedure with three separate hyperparameter searches.
\begin{itemize}
\item The minimal feature set (15 features per hero) only contains the current health, total gold, position, and the hero and tower proximity features. We selected these features because we think these are the most informative. The total gold describes how powerful the hero is, while the proximities describe who is in range to attack or help the hero.
\item The medium feature set (109 features per hero) contains everything except for the hero ID and the abilities features. These two feature categories occupy more than 60\% of the feature space. While a powerful ability can change the tide of the battle, making use of these features is difficult, since there are over four hundred unique abilities in the game.
\item The large feature set contains all features (287  features per hero).
\end{itemize}

\subsection{Training the Network}

The data set was divided prior to training into an 80\%/10\% training and validation split, with 10\% left over for testing. The training set consisting of 5848 matches, and the validation set consisting of 732 matches. After applying label balancing, the training set contained  57.6 million datapoints, while the validation set contained 7.2 million datapoints. 9.2\% of the labels were positive.

The trained network was evaluated on the test set containing 731 matches. There was no label balancing applied to the test set, such that the data remained a continuous match of data points representing the real distribution of data. The test data consisted of 14 million data points, 1.04\% percent of the labels were positive. 


Due to the sheer amount of data, it was impossible to load the entire data set into memory. Therefore the data was shuffled and saved into small files each containing 4000 data points. This resulted in roughly 16,200 files for the entire data set. For each training iteration, a batch of 128 data points was assembled from a random file.



For each feature set, we used a random search~\cite{bergstra2012random} based exploration to determine the best parameters to use for the following; the number of layers, number of neurons per layer, learning rate and batch size. A detailed description of the determined hyperparameters can be seen in Table~\ref{table:hyperparameters}.

\begin{table*}[ht!]
\centering
\small
\caption{Hyperparameters for each feature set. The architecture is given by lists, where the length of the lists are the number of layers, and the values are the number of neurons in each layer. See Fig.~\ref{fig:network} for an explanation of shared and final layers. }
\label{table:hyperparameters}
\begin{tabular}{@{}llll@{}}
\toprule
Hyperparameter & Minimal Features & Medium Features & All Features \\ 
 \midrule
Learning rate & 3.06 e-5 & 7.48 e-5 & 6.15 e-5 \\ 
Shared layers & {[}200, 100, 60, 20{]} & {[}256, 128, 64{]} & {[}256, 128, 64{]} \\ 
Final layers & {[}150, 75{]} & {[}1024, 512, 256, 128, 64, 32{]} & {[}1024, 512, 256, 128, 64, 32{]} \\ 
Batch Size &128&128&128\\
Optimizer & Adam&Adam&Adam\\
\bottomrule
\end{tabular}
\end{table*}



\begin{figure}
\centering
  \includegraphics[width=.99\linewidth]{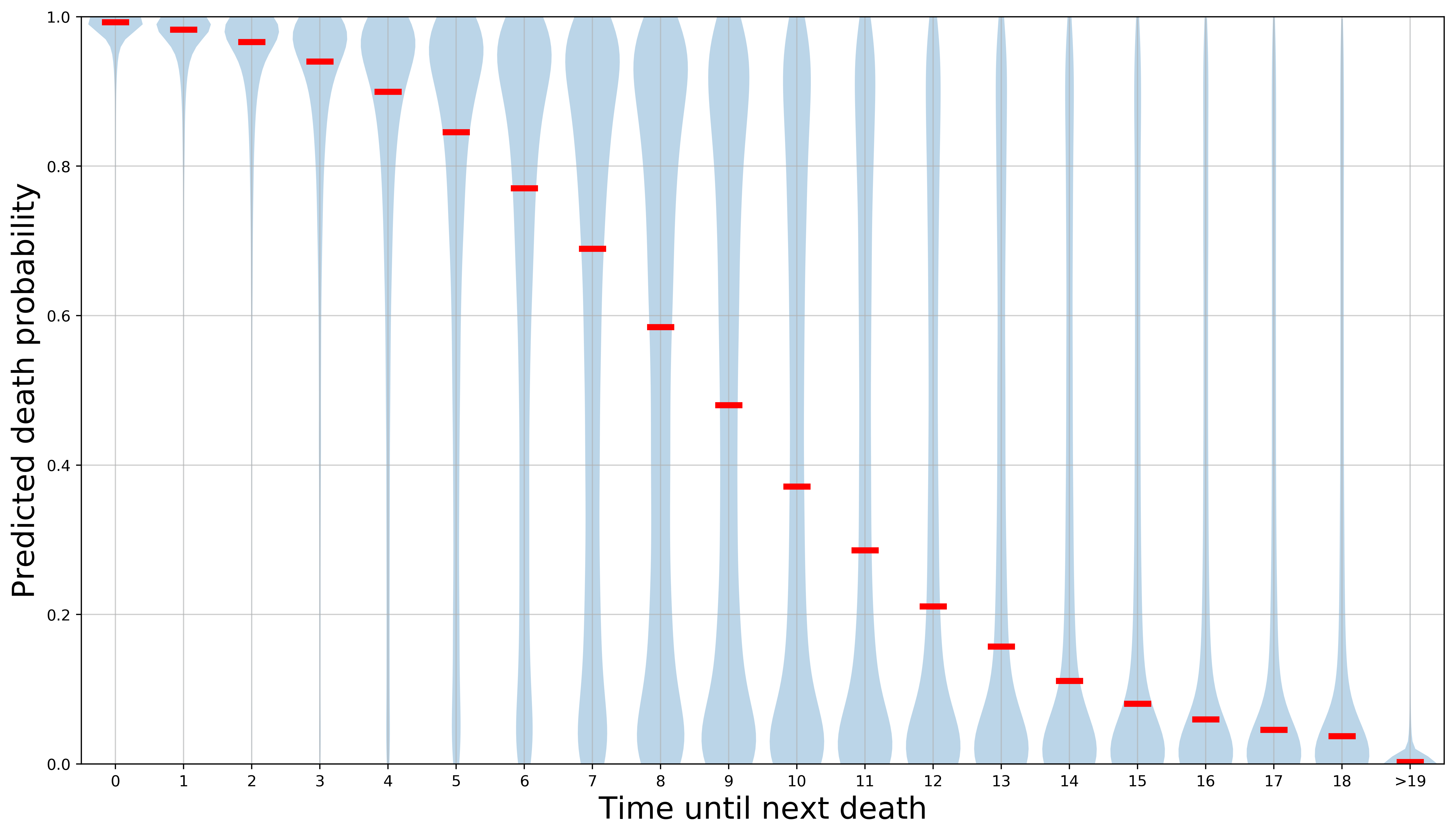}
\caption{Distribution of predictions for data points with different time until the player dies on the test data. The red line represents the median. The network is trained to predict whether a player will die within 5 seconds.}

\label{fig:results}
\end{figure}

\begin{figure}
\centering
  \includegraphics[width=1\linewidth]{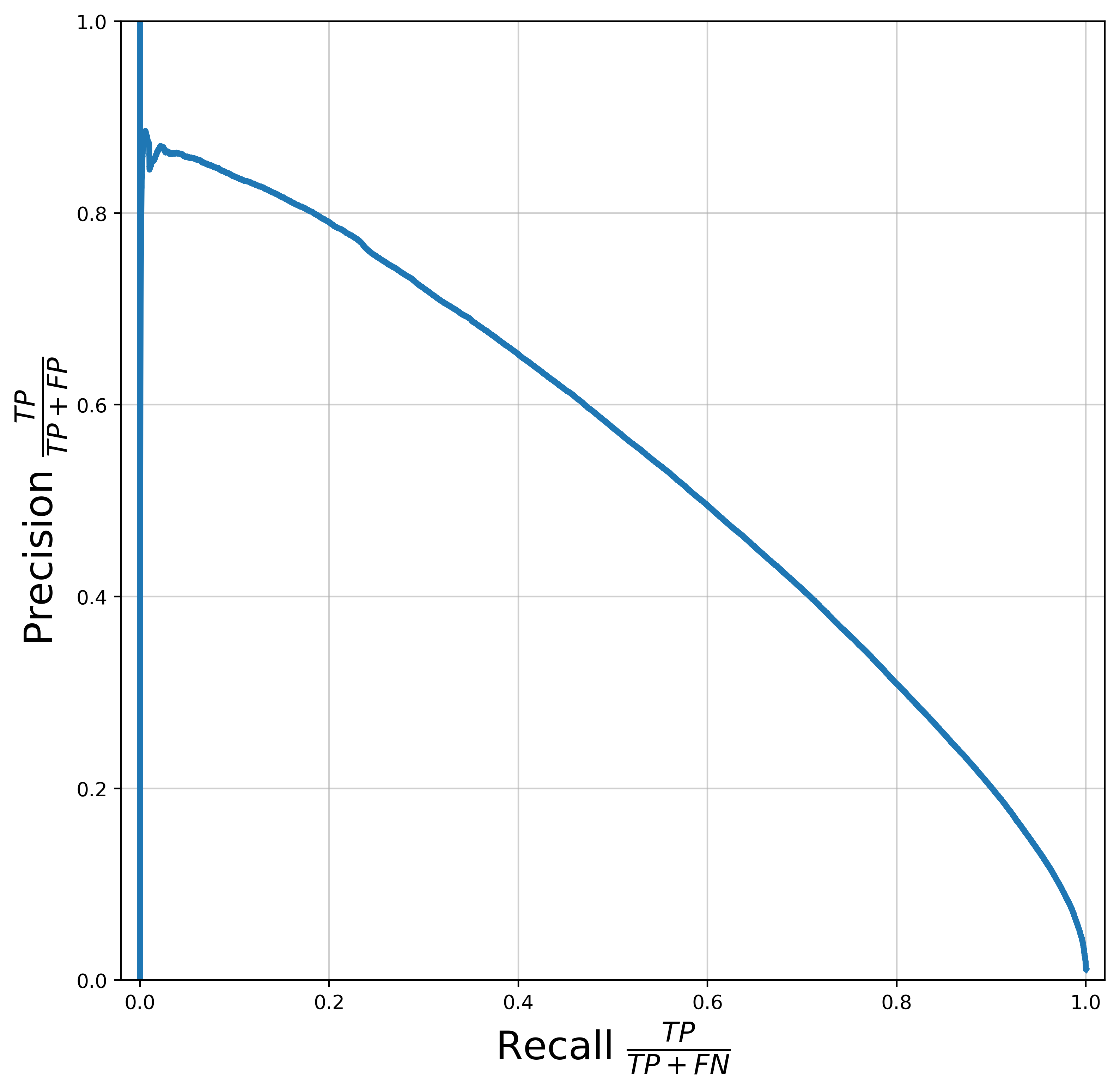}
\caption{Precision-Recall curve for the most accurate model, run against the entire test set.}
\label{fig:precisonRecall}
\end{figure}
\begin{table}
\small
\centering
\caption{Average precision (Area under the Precision-Recall curve) for feature sets for most accurate model. }

\label{table:results}
\begin{tabular}{@{}lr@{}}
    \toprule
    Feature Set & Average Precision  \\
    \midrule
    Minimal & 0.5001 \\
    Medium  & 0.5365  \\ 
    All & \textbf{0.5447}  \\
    \bottomrule
  \end{tabular}

\end{table}

\section{Results}\label{sec:Results}
\begin{figure*}
\centering

\begin{subfigure}[t]{0.99\textwidth}
   \includegraphics[width=1\linewidth]{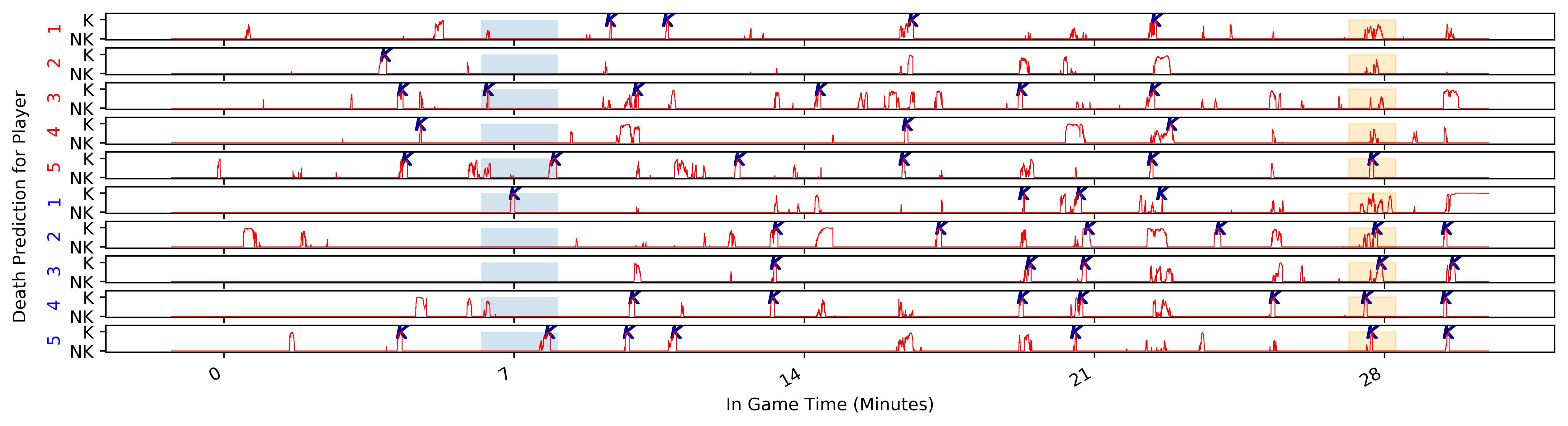}
    \caption{}
   \label{fig:gamePred}
\end{subfigure}

\begin{subfigure}[t]{0.49\textwidth}
   \includegraphics[width=1\linewidth]{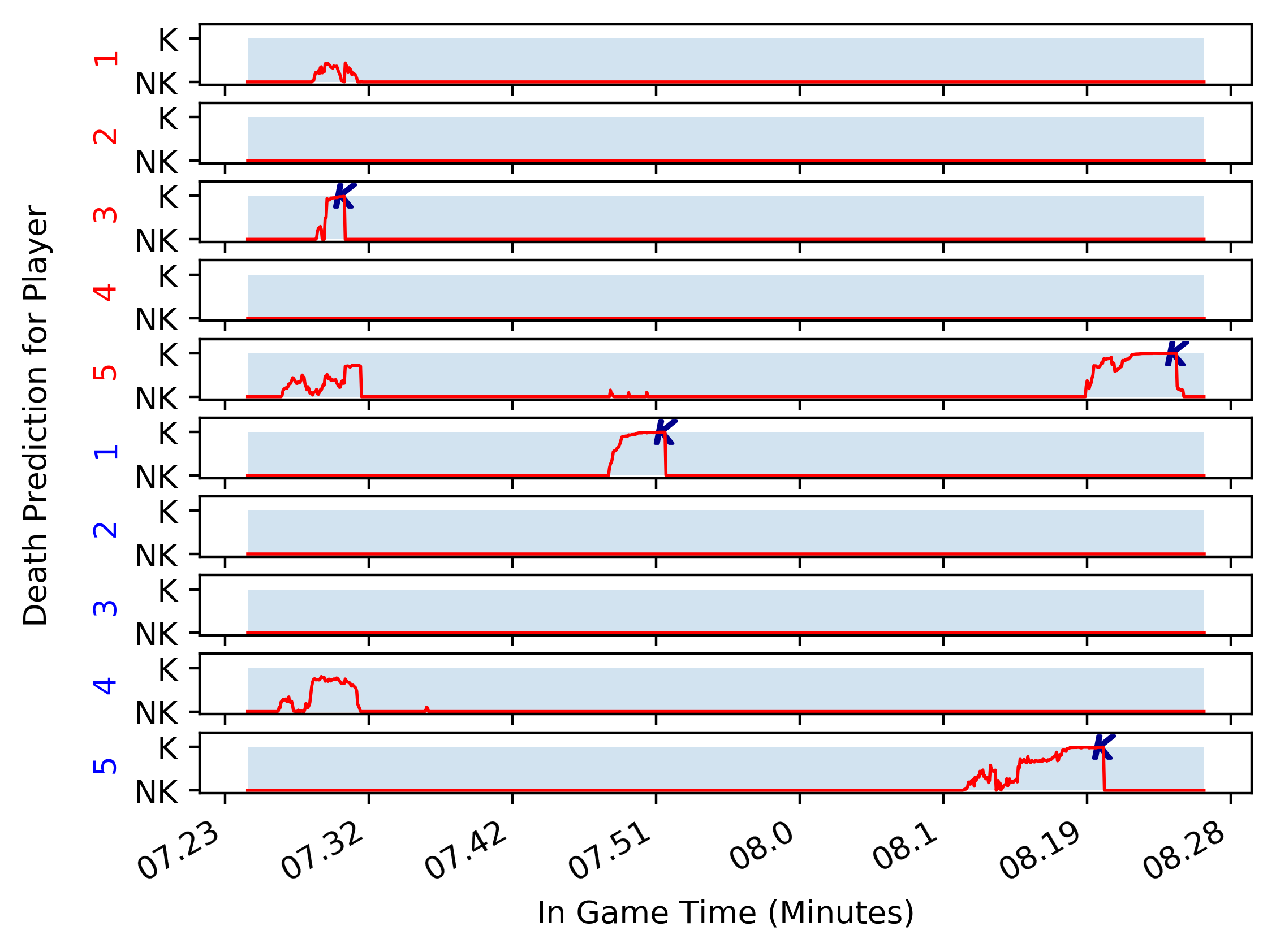}
   \caption{}
   \label{fig:gamePredSmallBlue} 
\end{subfigure}
\begin{subfigure}[t]{0.49\textwidth}
   \includegraphics[width=1\linewidth]{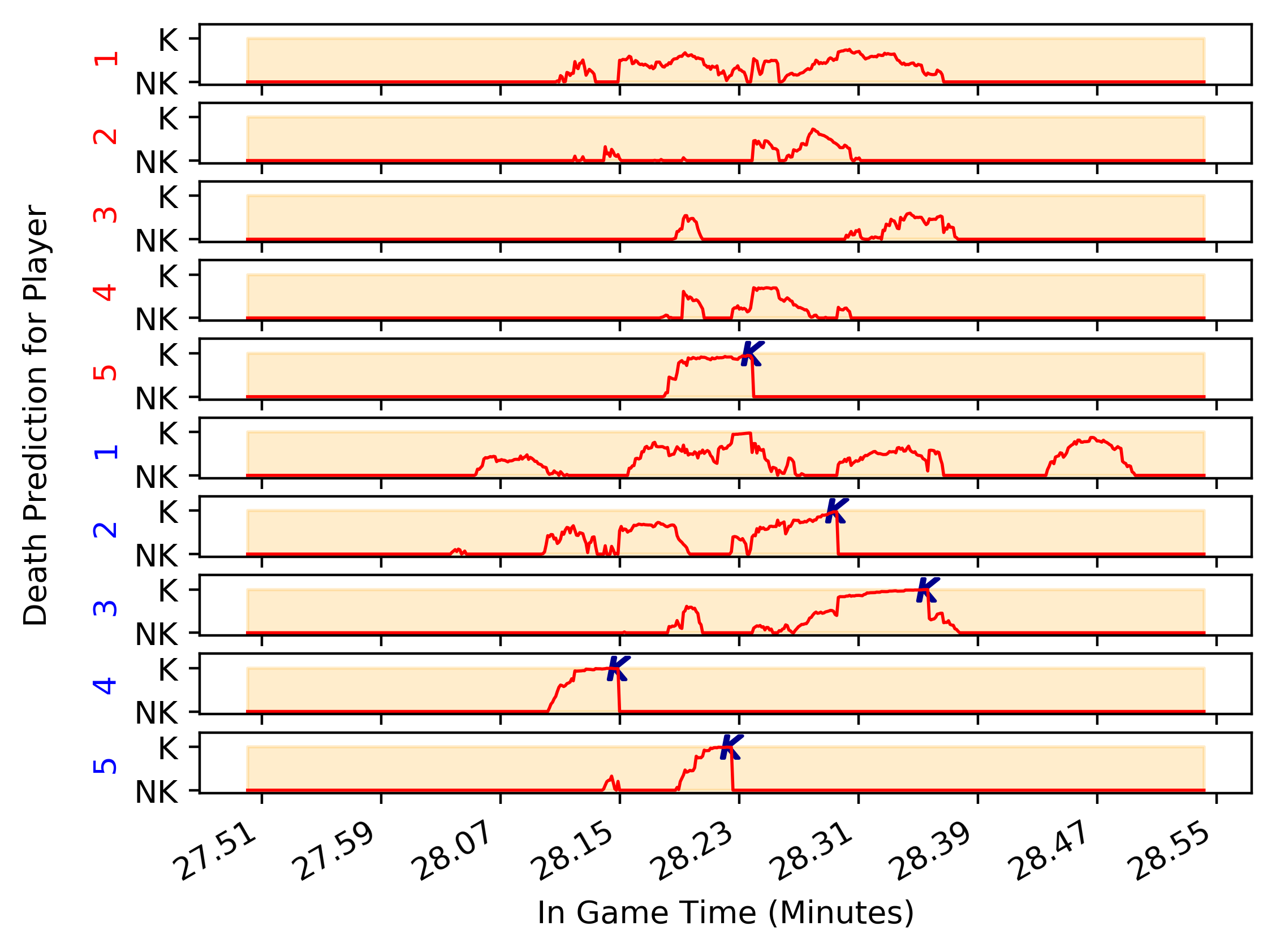}
   \caption{}
   \label{fig:gamePredSmallOrange} 
\end{subfigure}

\caption{(a) shows the results of a randomly sampled game (ID 2857921062) from the test set, run through our prediction model with our threshold set to 0.5. Visual indication of certainty of kills which are labelled as `K' with red probability line showing certainty of the death for each individual player, with teams annotated as red and blue.
(b) and (c) show smaller blown-up sections of the game, highlighting the two respective shaded regions in (a). }
\label{fig:wholeGamePlot}
\end{figure*}

Due to kills occurring rarely within a Dota 2 match, the ratio of positive labels in the test data is around 1\%. In such imbalanced data, accuracy is not a good measure of performance. Instead, we used the precision-recall curve, and the average precision, which is the area under the precision-recall curve.

Table~\ref{table:results} shows the performance of the best model on the test set for each feature set. 
The results indicate that the network is able to make use of the extra features. The performance advantage of the medium feature set to the minimal one is substantial. The advantage of the largest feature set over the medium one is more subtle. 

The best performing model achieved an average precision of 0.5447. We used this model for further analysis. The precision-recall curve of this model is shown in Fig.~\ref{fig:precisonRecall}. On the test data, a few of the highest predicted probability were actually false positives. These outliers are creating a narrow drop at the beginning of the precision-recall curve.  The precision-recall shows the dilemma between choosing a high threshold with high precision, and low recall, or choosing a low threshold with low precision and high recall. For example, at the threshold of 0.9 the model has 0.377 precision with 0.725 recall. By passing data in a continuous manner from a match we can construct a graph of prediction points on our test matches over the course of the entire match file. An example of a typical game predictions graph can be seen in Fig.~\ref{fig:gamePred}, in which we plot the actual labels of kills (K) from a game against the prediction output of the model. This can visualize how many true positives and false positives occur. 



It is also worth noting that even though we are testing for kills before 5 seconds, on some occasions kills would still be predicted before the 5-second mark, though to reduce any confusion in the calculations these would still be counted as mispredictions. Therefore if this tool would be used in a live game the accuracy of predictions could be higher than what is reported in this section. This is further discussed in Section~\ref{sec:discussion}. A generalised overview of prediction times from both the trained 0-5 seconds and an extended window to 20 seconds, and their probability outputs can be seen in Fig~\ref{fig:results}

To disprove any notion that the network relied heavily on certain feature correlations with death, we chose to analyse the feature of each player's health relevant to the prediction from the model. We analyzed an entire match to calculate the probability that these values were related. By plotting both of these variables on a scatter plot and conducting a Spearman rank correlation coefficient, giving a value of -0.09 with P~$< 0.001$. Effectively, heavily disproving any issues regarding the learning of the network regressing on one variable.

\section{Discussion}
\label{sec:discussion}


Dota 2's replay function provides the ability to replay matches exactly. While delving in to the replay for Fig.~\ref{fig:gamePred}'s match accompanied by our prediction tool, we could more closely monitor false predictions from the network, rather than witnessing them statistically as mis-classifying a kill. What we found indicated that although the network would be certain a player would die, a myriad of external factors would come into play. For example whether a player would commit to a kill while not having full information on the enemy team, or simply miscalculating their damage. A further study would need to be conducted to understand what is exactly happening within these moments, perhaps with professional domain assistance. What is certain is that at each spike of kill prediction an encounter between two or more heroes occurs.

To determine the usefulness of the model as a commentator assisting tool, the ability to predict interesting situations must be evaluated. Using standard performance metrics such as precision or recall are not satisfactory for this purpose, since different kinds of mispredictions have different impact on the usefulness of the model.
We can classify mispredictions into four categories:
\begin{enumerate}
\item False negatives. The hero will die, but the model failed to predict it.
\item False positives where hero is about to die, but not within 5 seconds. 
\item False positives where hero is not about to die, but professional players would evaluate the situation as dangerous. 
\item False positives where hero is not about to die, and professional players would evaluate the situation as not dangerous. 
\end{enumerate}
A model with a large number of category 1 errors can still provide assistance to a commentator, even though this assistance is provided less often.
On the other hand a large number of category 4 errors have a catastrophic effect on the usefulness of the model. If the model constantly alerts the commentator to uninteresting events, it becomes a hindrance instead of a help.

Category 2 and 3 mispredictions arguably make the model more useful and thus they are desirable.
Category 2 mispredictions can alert the commentator to fights where someone will die further ahead in time. A game situation when someone barely escaped can be just as interesting or entertaining as when someone died. The ability to predict such situations makes the model more useful, thus category 3 mispredictions are desirable. Unfortunately, to differentiate between category 3 and 4 mispredictions, human expert input is needed. For this reason evaluating the performance is difficult and requires further work. 



\section{Conclusion}\label{sec:conclusion}



In the above we have outlined our research on micro-predictions in the esports game Dota 2. Using a multilayered perceptron, trained on a vast array of features, that utilizes shared weights within the first half of the network, the ability to train a model that can accurately predict the death-state of every player within a live or recorded game of Dota 2, while efficiently dealing with the symmetric inputs of a MOBA data set. We further describe the process of pre-processing esports data from Dota 2 and how to replicate the experiments presented in their entirety, including the results of a large scale parameter exploration on network and feature architectures. Jointly, this will assist future experiments targeting micro-predictions into esports. The novel contribution of this paper is the exploration of deep learning models applied to Dota 2 at a micro level (or event level) as compared to match level. We provide a model for predicting the death of professional Dota 2 players with a high degree of accuracy within a 5 second window, with an example shown in Fig.~\ref{fig:gamePred}. Though we trained for a 5 second window of prediction, the model has learned inherent properties of features that constitute kills such that in certain situation the network can predict even prior to the 5 second window. The distribution of these probabilities can be seen in the violin plot in Fig.~\ref{fig:results}. 

The model can also be developed into a focus tool for game commentators that can direct their focus and storytelling upwards of 5 seconds before kills occur within the game. 




\bibliography{main}
\bibliographystyle{ieeetr}
\end{document}